\documentclass[sigconf, nonacm]{acmart}

\newcommand\vldbdoi{XX.XX/XXX.XX}

\newcommand\vldbavailabilityurl{URL_TO_YOUR_ARTIFACTS}

\usepackage{times}
\usepackage{latexsym}

\usepackage[T1]{fontenc}

\usepackage[utf8]{inputenc}

\usepackage{microtype}
\usepackage{graphicx}
\usepackage[capitalize]{cleveref}
\usepackage{array,multirow}

\usepackage{enumitem}

\usepackage{subcaption}
\usepackage[T1]{fontenc}
\usepackage{xcolor}
\usepackage{tabu}

\usepackage{amsmath}

\definecolor{flatdarkgray}{HTML}{7F8C8D}
\definecolor{flatgray}{HTML}{BDC3C7}
\definecolor{flatred}{HTML}{C0392B}
\definecolor{flatorange}{HTML}{D35400}
\definecolor{flatyellow}{HTML}{F39C12}
\definecolor{flatdenim}{HTML}{2C3E50}
\definecolor{flatpurple}{HTML}{8E44AD}
\definecolor{flatblue}{HTML}{2980B9}
\definecolor{flatgreen}{HTML}{27AE60}
\definecolor{flatcyan}{HTML}{16A085}
\definecolor{flatdarkgrayalt}{HTML}{95A5A6}
\definecolor{flatgrayalt}{HTML}{ECF0F1}
\definecolor{flatredalt}{HTML}{E74C3C}
\definecolor{flatorangealt}{HTML}{E67E22}
\definecolor{flatyellowalt}{HTML}{F1C40F}
\definecolor{flatdenimalt}{HTML}{34495E}
\definecolor{flatpurplealt}{HTML}{9B59B6}
\definecolor{flatbluealt}{HTML}{3498DB}
\definecolor{flatgreenalt}{HTML}{2ECC71}
\definecolor{flatcyanalt}{HTML}{1ABC9C}

\newtoggle{showtodos}
\togglefalse{showtodos}
\newcommand{\todo}[3]{\iftoggle{showtodos}{\textcolor{#1}{\textbf{TODO(#2): #3}}}{}}
\newcommand{\suggest}[2]{\iftoggle{showtodos}{\footnote{\textcolor{flatred}{#1's Suggestion: #2}}}{}}

\newcommand{\eat}[1]{}
\newcommand{\red}[1]{\textcolor{black}{#1}}

\newcommand{\yichaojoey}[1]{\todo{flatred}{yichaojoey}{#1}}

\newcommand{\procurement}{Supply Chain}
\newcommand{\banking}{Retail Finance}
\newcommand{\mortgage}{Tax Forms}

\begin{document}
\title{Radically Lower Data-Labeling Costs for Visually Rich Document Extraction Models}

\author{Yichao Zhou}
\affiliation{%
  \institution{Google}
  \streetaddress{}
  \city{Mountain View}
  \state{USA}
  \postcode{94043}
}
\email{yichaojoey@google.com}

\author{James B. Wendt}
\affiliation{%
  \institution{Google}
  \streetaddress{}
  \city{Mountain View}
  \state{USA}
  \postcode{94043}
}
\email{jwendt@google.com}

\author{Navneet Potti}
\affiliation{%
  \institution{Google}
  \streetaddress{}
  \city{Mountain View}
  \state{USA}
  \postcode{94043}
}
\email{navsan@google.com}

\author{Jing Xie}
\affiliation{%
  \institution{Google}
  \streetaddress{}
  \city{Mountain View}
  \state{USA}
  \postcode{94043}
}
\email{lucyxie@google.com}

\author{Sandeep Tata}
\affiliation{%
  \institution{Google}
  \streetaddress{}
  \city{Mountain View}
  \state{USA}
  \postcode{94043}
}
\email{tata@google.com}

\begin{abstract}
A key bottleneck in building automatic extraction models for visually rich documents like invoices is the cost of acquiring the several thousand high-quality labeled documents that are needed to train a model with acceptable accuracy. We propose {\em selective labeling} to simplify the labeling task to provide ``yes/no'' labels for candidate extractions predicted by a model trained on partially labeled documents. We combine this with a custom active learning strategy to find the predictions that the model is most uncertain about. We show through experiments on document types drawn from 3 different domains that selective labeling can reduce the cost of acquiring labeled data by $10\times$ with a negligible loss in accuracy.
\end{abstract}

\maketitle



\section{Introduction}
\label{section:introduction}

Visually rich documents such as invoices, receipts, paystubs, insurance statements, tax forms, etc. are pervasive in business workflows. The tedious and error-prone nature of these workflows has led to much recent research into machine learning methods for automatically extracting structured information from such documents \citep{lee2022formnet, garncarek2021lambert, xu-etal-2021-layoutlmv2, tata2021glean, Wu2018FonduerKB, Sarkhel2019VisualSF}. Given a target document type with an associated set of fields of interest, as well as a set of human-annotated training documents, these systems learn to automatically extract the values for these fields from documents with unseen layouts.

A critical hurdle in the development of high-quality extraction systems is the large cost of acquiring and annotating training documents belonging to the target types. 
The human annotators often require training not only on the use of the annotation tools but also on the definitions and semantics of the target document type. The annotation task can be tedious and cognitively taxing, requiring the annotator to identify and draw bounding boxes around dozens of target fields in each document. Not all the fields in the schema occur in all documents, leading to higher quality ground-truth annotations for the easier fields that occur frequently and lower quality annotations for infrequent fields, which are often missed. 

This data efficiency requirement has not gone unnoticed in the research literature on this topic. 
Pre-training on large unlabeled document corpora \citep{xu2020layoutlm,xu-etal-2021-layoutlmv2},  transfer learning from an out-of-domain labeled corpus \citep{torrey2010transfer, nguyen2019transfer}, and data-programming~\citep{Ratner2017SnorkelRT, arxiv.2202.05433} have all proven to be useful techniques in reducing the amount of training data required to get accurate models. However, even with these techniques, empirical evidence suggests that performing well on a new target document type still requires thousands of annotated documents, amounting to hundreds of hours of human labor \citep{zhang2021visual}. Automating document-heavy business workflows in domains like procurement, banking, insurance, mortgage, etc. requires scaling to extraction models for hundreds of different document types.

The cost of acquiring high quality labeled data for hundreds of document types is prohibitively expensive and is currently a key bottleneck. We could apply active learning strategies to select a few but informative documents for human review~\citep{settles2009active}, however the cost-reducing effect of this approach is limited, as it requires annotating the span in every selected document for every field. Many of these annotations are repetitive, and often not very informative if a model can already extract those fields easily. In fact, our initial experiments with a document-level active learning approach yielded modest results that cut down the number of documents required to get to the same level of quality as random selection by approximately 20\%.
In this paper, we propose a technique called {\em selective labeling} that reduces this cost by $10\times$. The key insight is to combine two ideas: First, we redefine and simplify the task performed by the human annotators -- rather than labeling every target field in every document by drawing bounding boxes around their values, we ask them to simply verify whether a proposed bounding box is correct. This binary ``yes/no'' annotation task is faster and imposes a lighter cognitive burden on the annotator \citep{BoimGMNPT12,cloudresearchblog,ganchev2007semi,skeppstedt2017pal}. Second, we adapt existing active learning strategies to select the examples (i.e., candidate extraction spans) that the model is most uncertain in each round to annotate. In other words, we consider active-learning at the (document, field)-pair granularity rather than at the document level granularity and choosing a modeling approach that can easily deal with the complexity resulting from the partially labeled documents this approach produces.

We find that relying on a simple uncertainty metric, such as the distance between prediction scores and the middle point between the target labels (e.g., 0.5), is sufficient for selecting informative candidate extraction spans to annotate. We further propose new methods to increase diversity in the selection pool by reallocating the annotation budget to encourage selection of more infrequent fields. This is accomplished by calibrating the highly imbalanced prediction scores at the field level and limiting the number of candidates of each field to be reviewed in each document.

\begin{figure}[t]
\includegraphics[width=.85\linewidth]{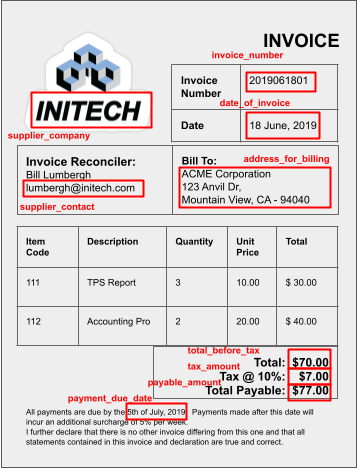}
\centering
\caption{A classic annotation task. Even labeling 9 fields in this toy invoice imposes a heavy cognitive burden on the annotator, while real-world documents are significantly more complicated.}
\label{fig:classic_annotation_task}
\end{figure}

\begin{figure}[t]
\includegraphics[width=.85\linewidth]{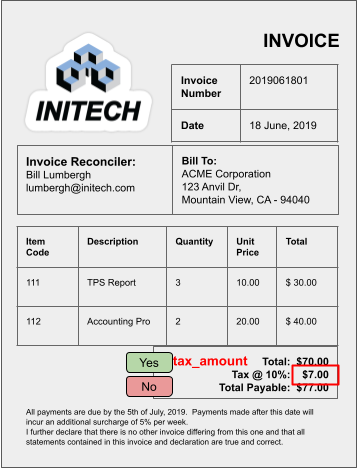}
\centering
\caption{A ``yes/no'' annotation task. Presenting a proposed span and asking the annotator to accept or reject the label is simpler, quicker, and less prone to errors.}
\label{fig:yesnotask}
\end{figure}

We interleave rounds of such human annotation with training a model that is capable of consuming partially labeled documents. In combination, our proposed approach dramatically improves the efficiency of the annotation workflow for this extraction task. In fact, through experiments on document types drawn from multiple domains, we show that selective labeling allows us to build models with $10\times$ lower annotation cost while achieving nearly the same accuracy as a model trained on several thousand labeled documents. Note that our goal in this paper is {\em not} to advance the state-of-the-art in active learning, {\em nor} to propose a more data-efficient model for extraction from layout-heavy documents. Our main contribution is that we demonstrate that a novel combination of an existing active-learning strategy with an existing extraction model can be used to {\em dramatically cut down the primary bottleneck} in developing extraction models for visually rich documents.

\section{Background}\label{sec:background}

We first describe how a typical annotation task is set up to acquire labeled documents. We point out two major deficiencies with this approach before outlining an alternative that takes advantage of the characteristics of this domain. We then describe the assumptions underlying our approach.
\eat{\suggest{nav}{I rewrote most of this section to make it easier to skim. Feel free to revert any undesired changes. Consider renaming this section to "Annotation Workflow" or something similar. I also removed the intro paragraph here. If we bring it back, we should rewrite it to summarize everything in this section; it currently leaves out the proposed workflow and the assumptions.}}

\subsection{Annotation Workflow}\label{annotation_workflow}
\eat{\suggest{nav}{This paragraph doesn't introduce all the things we plan to talk about in this section. In particular, it doesn't mention that we are also going to talk about the proposed annotation workflow and the assumptions underlying our approach. I think the right fix is to break this section down or split it.}}

\subsubsection{Classic Annotation Workflow} \label{sec:classic_annotation_workflow}

Given a document type for which we want to learn an extraction model, we begin by listing out the fields that we want to extract, along with human-readable descriptions, viz., ``labeling instructions''. We provide these instructions to human annotators and present them with various document images to label. The classic annotation task is to draw a bounding box around each instance of any of the target fields and label it with the corresponding field name (Figure~\ref{fig:classic_annotation_task}). Typical document types like invoices and paystubs have dozens of fields, and each document may contain multiple pages.

The high cognitive burden of the classic annotation workflow leads to two major drawbacks. First, it makes training data collection extremely expensive. In one annotation task for paystub-like documents with 25 target fields, the average time to label each document was about 6 minutes. Scaling this to hundreds of document types with thousands of documents each would be prohibitively expensive. Second, the resulting annotation quality is often quite poor. We have observed systematic errors such as missing labels for fields that occur infrequently in the documents or for instances that are in the bottom third of the page. To obtain acceptable training and test data quality, each document must be labeled multiple times, further exacerbating the annotation cost issue.

\subsubsection{Proposed Annotation Workflow} \label{sec:proposed_annotation_workflow}

We propose the following alternative to the classic annotation workflow:

\begin{enumerate}[itemsep=0pt]
    \item We speed up labeling throughput by simplifying the task: rather than drawing bounding boxes, we ask human annotators to accept or reject a candidate extraction. Figure~\ref{fig:yesnotask} illustrates how much easier this ``yes/no'' task is compared to the classic one in Figure~\ref{fig:classic_annotation_task}.
    
    \item We further cut down annotation cost by only labeling a subset of documents and only a subset of fields in each document. 
    
    \item We use a model trained on partially labeled documents to propose the candidate extraction spans for labeling. This allows us to interleave model training and labeling so that the model keeps improving as more labels are collected. 
    
    \item We use a customized active learning strategy to identify the most useful labels to collect, viz., the candidate extraction spans about which the model is most uncertain. In successive labeling rounds, we focus our labeling budget on the fields that the model has not yet learned to extract well, such as the more infrequent ones. 
\end{enumerate}


In Section~\ref{sec:results}, we show empirical evidence that this improved workflow allows us to get to nearly the same quality as a model trained on 10k docs by spending an {\em order-of-magnitude less} on data-labeling. Note that naively switching the labeling task to the ``yes/no'' approach does not cut down the labeling cost -- if we were to highlight every span that might potentially be an amount and present an ``Is this the tax\_amount?'' question like in Figure~\ref{fig:yesnotask}, with the dozens of numbers that are typically present in an invoice, this workflow will be {\em much more} expensive than the classic one. A key insight we contribute is that a model trained on a modest amount of data can be used to determine a highly effective subset of ``yes/no'' questions to ask.

\subsection{Assumptions} \label{sec:assumptions}
We make the following four assumptions about the problem setting: 
\begin{enumerate}[itemsep=0pt]
    \item We assume access to a pool of unlabeled documents. This is a natural assumption in any work on managing cost of acquiring labeled training data.
    
    \item We assume the extraction model can be trained on partially labeled documents.
    
    \item We assume the model can generate candidate spans for each field and a measure of uncertainty -- this is used to decide the set of ``yes/no'' questions to present to the annotator. 
    
    \item The analysis in this paper uses empirical measurements for labeling tasks on documents with roughly 25 fields to model the costs of the traditional approach (6 minutes per document) and the proposed approach (10 seconds per ``yes/no'' question~\cite{cloudresearchblog}). For more complex documents the difference in the two costs may be significantly higher.
\end{enumerate}

\begin{figure}[t]
\includegraphics[width=.95\linewidth]{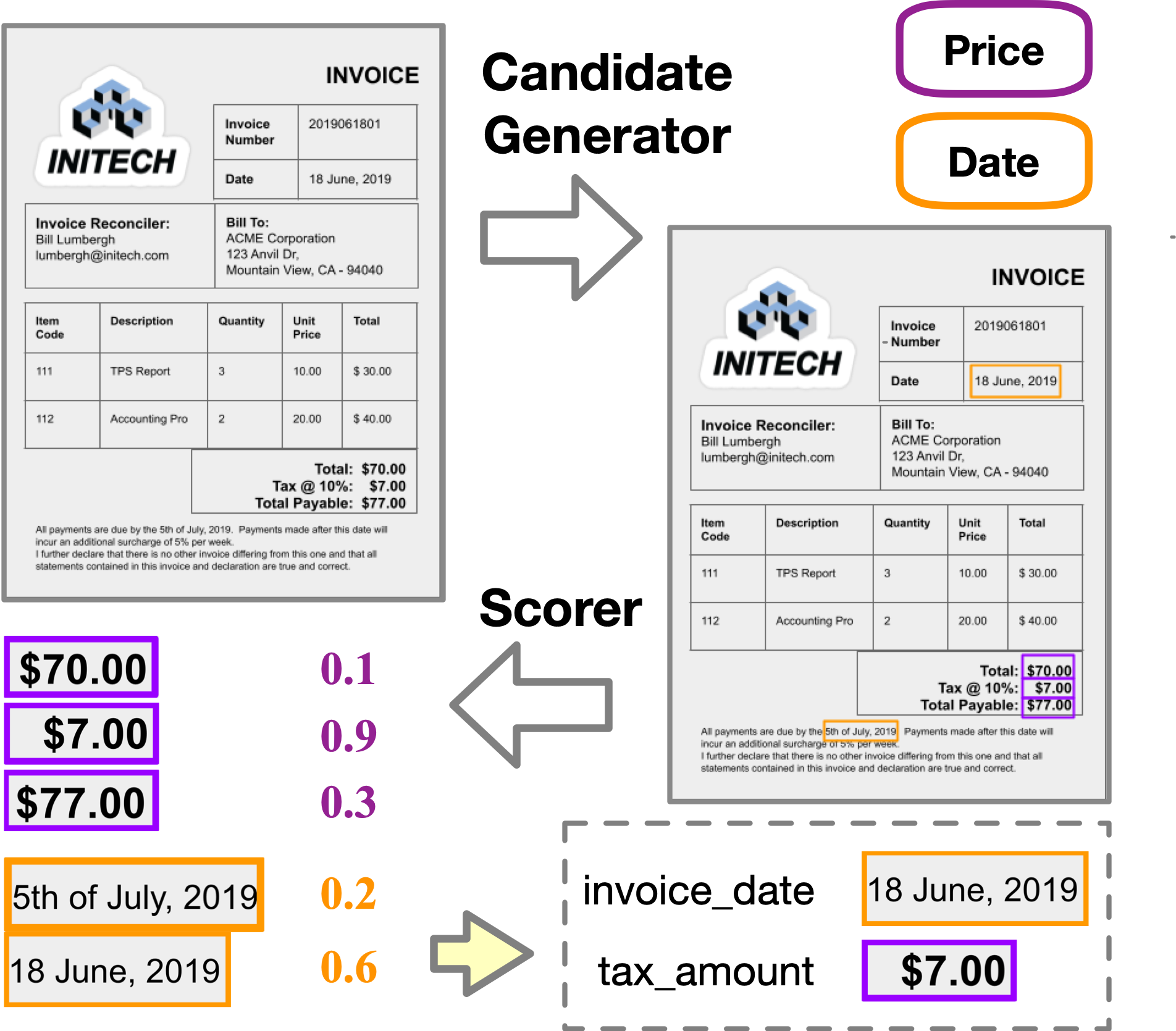}
\centering
\caption{Architecture of the candidate generator and scorer of our document extraction model. The scoring is done using a neural network model trained as a binary classifier.}
\label{fig:glean}
\end{figure}

Throughout this work, we use an extraction system similar to the architecture described in \citep{majumder2020representation}. As shown in Figure~\ref{fig:glean}, this architecture consists of two stages: candidate generation and candidate classification. In the first stage, we generate candidates for each field according to the type associated with that field. For example, the candidates generated for the  \textit{date of invoice} field would be the set of all dates in that invoice. The candidate generators for field types like dates, prices, numbers, addresses, etc. are built using off-the-shelf, domain agnostic, high-recall text annotation libraries. 
\red{The recall of candidate generation varies across fields, e.g. high in dates and prices while relatively low in addresses and names. Having a candidate generator with low recall indeed limits the recall of the final extractions for that field.}
In the second stage, we score each candidate's likelihood of being the correct extraction span for the document and field it belongs to. This scoring is done using a neural network model trained as a binary classifier. The highest-scoring candidate for a given document and field is predicted as the extraction output for the document and field if it exceeds a certain field-specific threshold.

The ability to train on partially labeled documents is trivially true for this modeling approach since it employs a binary classifier trained on the labeled candidates. This should be relatively straightforward for sequence labeling approaches, such as ~\cite{xu-etal-2021-layoutlmv2}, as well. Identifying a potential span in the document to present as a ``yes/no'' question to an annotator is an exercise in ranking the candidates for each field.  We expect that sequence labeling approaches can be adapted to satisfy this requirement, e.g., by using beam search to decode the top few sequence labels. However, this is likely more complex than the aforementioned approach, and we leave this as an exercise for future work.

\section{Selective Labeling Methodology}
\label{section:techniques}
\begin{figure}[t]
\includegraphics[width=.85\linewidth]{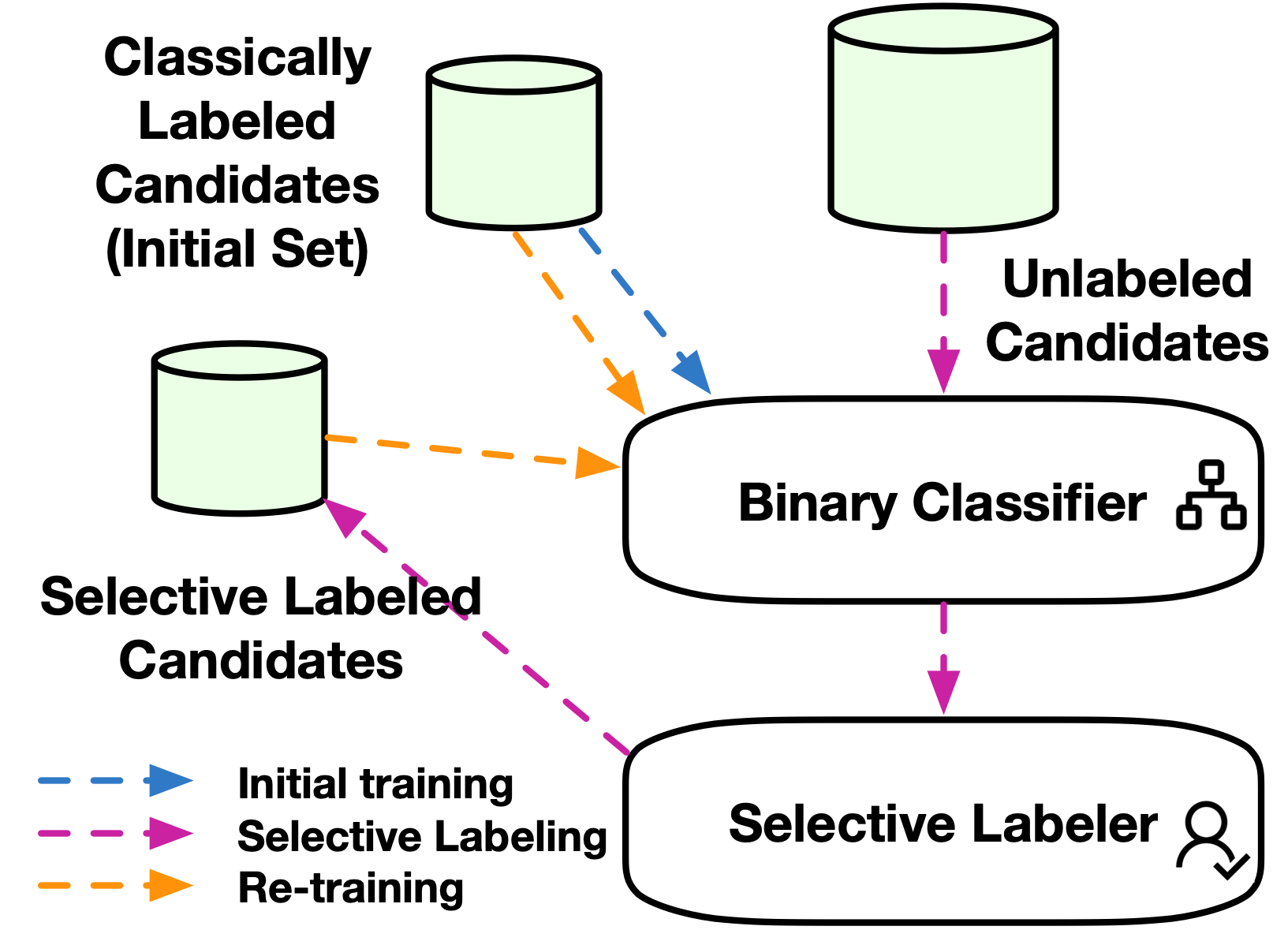}
\centering
\caption{The model training pipeline starts by inital training (blue) the binary classifier using the small classically labeled dataset. We then selectively label (purple) a fixed number of candidates according to the budget, which are then used to re-train (orange) the model together with the initial dataset. 
}
\label{fig:pipeline}
\end{figure}
We first provide an overview of the selective labeling framework before describing various uncertainty measures and ways to deal with the unique characteristics of this setting, such as varying difficulty for different fields.

\subsection{Overview} \label{sec:selective_labeling_overview}

Figure~\ref{fig:pipeline} provides a visual overview of our selective labeling workflow. We assume a scenario in which a corpus of several thousand unlabeled documents $U^d$  belonging to the target document type is available and we can request annotations from a labeler for every unlabeled document $d_i \in U^d$ which consists of a set of candidates $\{c_0^{d_i}, c_1^{d_i}, c_2^{d_i}, ..., c_n^{d_i}\}$. 
We begin by fully labeling a small randomly sampled subset of documents $S^d \subseteq U^d$, say 50-250 documents, using the classic annotation workflow. 
We learn an initial document extraction model $f(x|S^c)$, where $S^c$ represents the candidate set contained in $S^d$ and we mark all the remaining unlabeled candidates in $U^d\backslash S^d$ as $U^c$. 
Our labeling workflow proceeds in rounds. In each round $j$, the model is used to select candidates $S^c_j$ from $U^c$ and have them reviewed by human annotators. The annotators answer a ``yes/no'' question (see Figure~\ref{fig:yesnotask}) either accepting or rejecting this proposed label. As a result, $S^c = S^c \cup S_j^c$ and $U^c = U^c \backslash S_j^c$, meaning the newly labeled examples are merged into the training set and removed from the unlabeled set. The model is retrained on $S^c$ in each round and we repeat this iterative labeling-and-training procedure until we exhaust our annotation budget or reach our target F1 score. 




The efficacy of this workflow clearly depends on the procedure we use to select candidate spans for human annotation. Based on the fundamental insight underlying much active learning literature, we select the candidates that the model is \emph{most uncertain} about. In the remainder of this section, we describe how we adapt standard active learning strategies to a document extraction setting.

\subsection{Measuring Uncertainty}
\label{sec:selective_labeling_uncertainty}
There are a number of metrics we can use to quantify a model's prediction uncertainty~\citep{lewis1994sequential,ko1995exact}. In this work, we explored two types of uncertainty metrics.

\noindent\textbf{Score distance.} This method assigns a metric to each candidate based on the distance that the score is from some threshold \cite{li2006confidence}. More formally, the uncertainty is $1 - |score - threshold|$. For example, if the threshold is 0.5, this suggests that the model is most uncertain of its predictions of scores close to 0.5, in either direction. \red{This approach can also be interpreted as an entropy-based uncertainty method, where we find an optimal candidate $x^*$ so that $x^* = \underset{x}{\arg\max} -\Sigma_i P(y_i|x)\text{log}(P(y_i|x))$. In our binary classification setting, $y_i=\{0,1\}$ and candidates with scores closer to 0.5 results in larger entropy.}



\noindent\textbf{Score variance.} This method performs inference on a candidate multiple times with the dropout layer enabled and assigns the uncertainty metric as the variance of the scores \cite{gal2016dropout, kirsch2019batchbald,ostapuk2019activelink}. An alternative method trains multiple models independently from one another and assigns the uncertainty metric as the variance of the scores across all models~\citep{seung1992query}. Note that empirically, we observed this yields near identical results as the dropout-based approach, so we only present findings for the latter.

\subsubsection{Score Calibration}
\label{sec:calibration}
Our model's predicted scores tend to be un-calibrated \red{(as is very typical of neural networks \cite{guo2017calibration})}, particularly in initial rounds and for infrequent fields due to training data scarcity.
We calibrate scores in such a way that picking a candidate with a calibrated score of, say, $0.6$ yields a $60\%$ probability that it has a positive label~\citep{guo2017calibration}. \eat{We compute calibration curves using the labeled training dataset by bucketing the candidates based on score.}\red{We use a modified version of histogram binning \cite{zadrozny2001obtaining} and IsoRegC \cite{zadrozny2002transforming} to accommodate the highly non-uniform distribution of scores. However, any calibration method that deals with skewed distributions is suitable.}
Note that we recompute the calibration curves for the new model after every round of selective labeling.

\eat{There are two interesting design choices we made in this process, both of which are made based on our knowledge of the score distribution. (1) The vast majority ($>90\%$) of our candidates are negative and most of them have very low scores ($<10^{-3}$), while the region of interest to us when calibrating the scores is the rest ($[10^{-3}, 1]$). In calculating bin edges, we exclude all candidates with scores that are smaller than a threshold ($10^{-3}$). All the scores below this threshold are placed in the first bin ($[0, 10^{-3})$). Since the vast majority of candidates get excluded by this filter, the remaining bins have a much higher resolution. (2) We use equal-frequency bins rather than equal-width bins because of the highly non-uniform distribution of scores, even within the score region of interest -- in other words, each bin has roughly the same number of scores, except the first bin. 

Once binned, calibration curves are computed for each field by interpolating between the curves \emph{prevalence} (i.e., the proportion of candidates in each score bin that are positive) and the median scores for all the score bins.}  

By calibrating the scores, threshold selection becomes much more intuitive for the score-based uncertainty metric. For example, if we specify a threshold of $0.5$, we expect that to mean we will select candidates for which the model has a $50\%$ chance of classifying correctly \textit{across all fields}. \eat{Without calibration, it is unclear if a single global threshold applies equally well across all fields. \suggest{nav}{I added a sentence at the beginning of this subsection for better flow. I think that sentence covers this point.}}

\begin{table}[t]
    \centering
    \resizebox{\linewidth}{!}{
    \begin{tabular}{c|c|c|c|c}
    \hline
        Domain & \# Fields & Splits & \# Docs & \# Candidates \\
        \hline
        \hline
        \multirow{5}{*}{\procurement} & \multirow{5}{*}{18} & Initial-50 & 50 & 11.8K \\
        & & Initial-100 & 100 & 24.5K \\
        & & Initial-250 & 250 & 58.7K \\
        & & Test & 5,019 & 1.2M \\
        & & Hidden-label & 10,000 & 2.4M \\
        \hline
        \multirow{3}{*}{\banking} & \multirow{3}{*}{11} & Initial-100 & 100 & 76.0K \\
        & & Test & 849 & 1.2M \\
        & & Hidden-label & 4,000 & 5.6M \\ 
        \hline
        \multirow{3}{*}{\mortgage} & \multirow{3}{*}{24} & Initial-100 & 100 & 13.4K \\
        & & Test & 1,498 & 1.0M \\
        & & Hidden-label & 7,500 & 5.1M \\ 
        \hline
    \end{tabular}
    }
    \vspace{5pt}
    \caption{Statistics of datasets in three domains.}
    \label{tab:datasets}
    \vspace{-20pt}
\end{table}

\subsection{Sampling Candidates} \label{sec:selective_labeling_sampling}
Once the uncertainty metric is calculated for each candidate in the unlabeled set, the next step is to select a subset of those candidates for human review. The most obvious method is to select the top-$k$ candidates, thereby selecting the candidates for which the model is most uncertain. In practice, this can lead to sub-optimal results when the model finds many examples for which it is uncertain but may in fact be very similar to one another. The most common approach to break out of this trap is to introduce some notion of diversity in the sampling methodology~\citep{gao2020consistency,ishii2002control}.
\vspace{5pt}

\noindent\textbf{Combining Top-$k$ and Random Sampling.}
A common method is to reallocate the $k$ budget in each round so that a portion of that budget goes towards the top candidates by uncertainty (ensuring we get labels for the most uncertain candidates) and the remaining budget goes towards a random sample of candidates from the unlabeled set (ensuring that some amount of diversity is included in each round). One approach is to select the top-$k'$ candidates by the uncertainty metric, where $k' < k$, and then randomly sample $k-k'$ candidates from the remaining unlabeled dataset. A second approach is to randomly sample $k$ candidates from a pool of top-$n$ candidates, where $n > k$. We found in practice that these two methods yield nearly identical results, so we only present findings for the first.
\vspace{5pt}

\noindent\textbf{Capping Candidates for Each Document and Field.}
An important observation we make about the extraction problem is the following: While a given field typically has multiple candidates in every document, there are usually few positives per document compared to the number of negatives. For example, there are usually many dates in an invoice, and typically only one of them is the \textit{date of invoice}. The uncertainty metrics we defined in Section~\ref{sec:selective_labeling_uncertainty} do not take into account this relationship between labels.
We leverage this intuition to increase sample diversity by capping the number of candidates selected from the same document and field. After ordering the candidates by the chosen uncertainty metric, if we were to simply select the top-$k$ candidates, we might end up selecting too many candidates for the same document and field. Instead, we select at most $m$ candidates for each document and field, $m$ being a tunable hyperparameter we can adjust on a per-field basis. This ensures that we spread the annotation budget over more documents and fields. 

\subsection{Automatically Inferring Negatives}
After candidates have been selected and labeled, we merge the newly-labeled candidates into our training set. At this point, there is another opportunity to draw additional value from the unlabeled corpus by utilizing the structure of the extraction problem, in particular, for fields that are defined in the domain's schema to only have a single value per document (such as a document identifier, statement date, amount due, etc.). The key insight here is that when a positive label is revealed via selective labeling, we can infer negative labels for some remaining candidates in the document. If the schema indicates that a particular field is non-repeating, we can automatically infer that all of that field's remaining candidates in the document are negative.

\section{Experiment Setup}
\label{sec:setup}
To evaluate the performance of our proposed methods, we use datasets belonging to three different domains, summarized in Table~\ref{tab:datasets}. The number of fields varies across domains, e.g., the \textit{\mortgage} dataset has more than twice the fields as the \textit{\banking} dataset.
We use hidden-label datasets instead of real unlabeled datasets and simulate the labeling procedure by revealing the labels of the candidates from the hidden-label datasets.

\begin{table}[t]
    \centering
    \resizebox{.95\linewidth}{!}{
    \begin{tabu}{l|c|c}
    \hline
    Hyperparameter & Range explored & Best performer \\
        \hline
        \hline
        learning rate & 0.0001-0.1 & 0.001 \\
        dropout rate & 0.1-0.5 & 0.1 \\
        batch size & \{64, 128, 256\} & 128 \\
        top-$k'$ uncertain candidates & 0.7-1.0$k$ & 0.9$k$ \\
        $m$ candidates each field doc & 1-3 & 1 \\
        
        \hline
    \end{tabu}
    }
    \vspace{5pt}
    \caption{Hyperparameter selection.}
    \label{tab:parameter}
    \vspace{-20pt}
\end{table}

Recall from Section~\ref{sec:background} that we employ two annotation methods: \red{the classic annotation method (6 minutes per document), which is always applied to the initial training set, and the proposed ``yes/no'' method (10 seconds per candidate), which is applied during the selective labeling procedure on the unlabeled dataset.\footnote{\red{Targeting 10\% of the cost to fully label the unlabeled dataset via the classic annotation method, translates to selectively labeling 36k, 14k, and 27k ``yes/no'' questions for \textit{\procurement}, \textit{\banking}, and \textit{\mortgage} domains according to the estimation of same amount of annotation hours. If we bootstrap the model using the classic annotation workflow on a small number of documents, we simply subtract that cost from the budget for selective annotation.}}}
To explore how the size of the initial labeled dataset impacts our methods, we create three initial splits for the \textit{\procurement} domain with $50$, $100$, and $250$ documents. 
In all of our experiments, we split the train set into 80-20 training-validation sets. The validation set is used to pick the best model by AUC-ROC, and we use the test split to report the performance metrics. We train using the Rectified Adam~\citep{liu2019variance} optimizer and measure AUC-ROC on the validation set to decide whether to trigger early stopping after 3 epochs of no improvement. The binary classifier has 330k parameters and each set of experiments trained within 4 hours on a NVIDIA Tesla P100 GPU. \red{We apply grid search to tune the hyperparameters. The most performant hyperparameter values are listed in Table~\ref{tab:parameter}. Finally, we evaluate our methods by measuring the overall extraction system's performance on the test set using the maximum F1 averaged across all fields, denoted as ``Average E2E Max F1'' in \citep{majumder2020representation}. Every reported F1 score is further averaged over 10 independent runs to account for variability. All F1 scores are generated by comparing the extractions with the ground truth. If a field has a poor candidate generator, its final recall can obviously not exceed the recall of the candidate generator. }


\section{Results} \label{sec:results}
In this section, we present the overall performance of our best selective labeling strategy on three domains, a comparison of the different selection metrics, sampling methodologies, and how the number of rounds of selective labeling affects performance. We perform an ablation study to understand the effectiveness of our proposed diversity techniques, and finally demonstrate how performance varies with the size of the initial labeled dataset.

\begin{figure}[t]
\includegraphics[width=.9\linewidth,trim={0 0 0 2cm},clip]{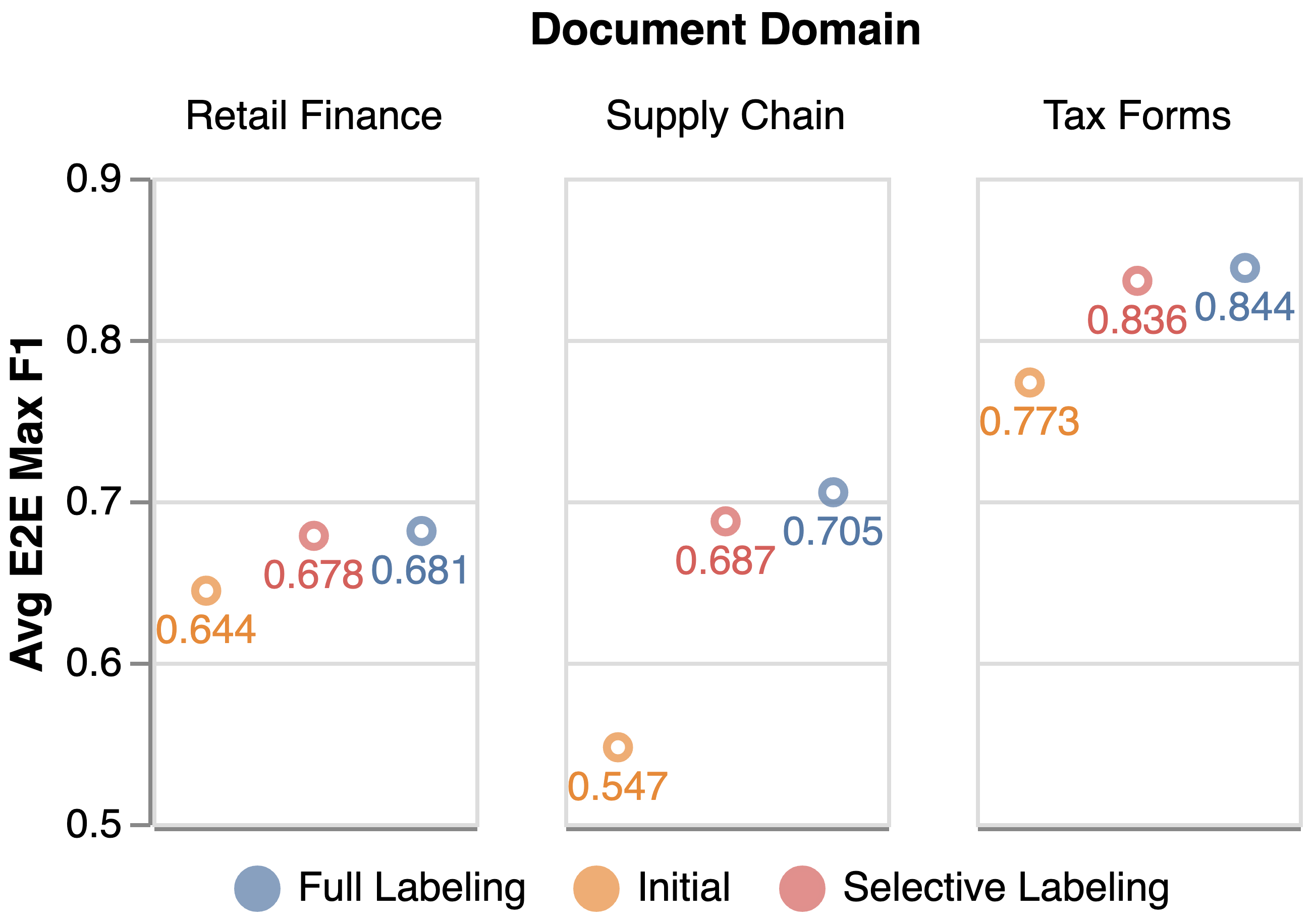}
\centering
\caption{Best performing Selective Labeling as compared to Initial which is trained on just 100 documents and Full Labeling in which the hidden-label dataset (used in Selective Labeling) is fully used in training.
}
\label{fig:domains}
\end{figure}

\begin{figure*}[!t]
    \centering
    \begin{subfigure}[t]{.30\linewidth}
        \includegraphics[width=\linewidth]{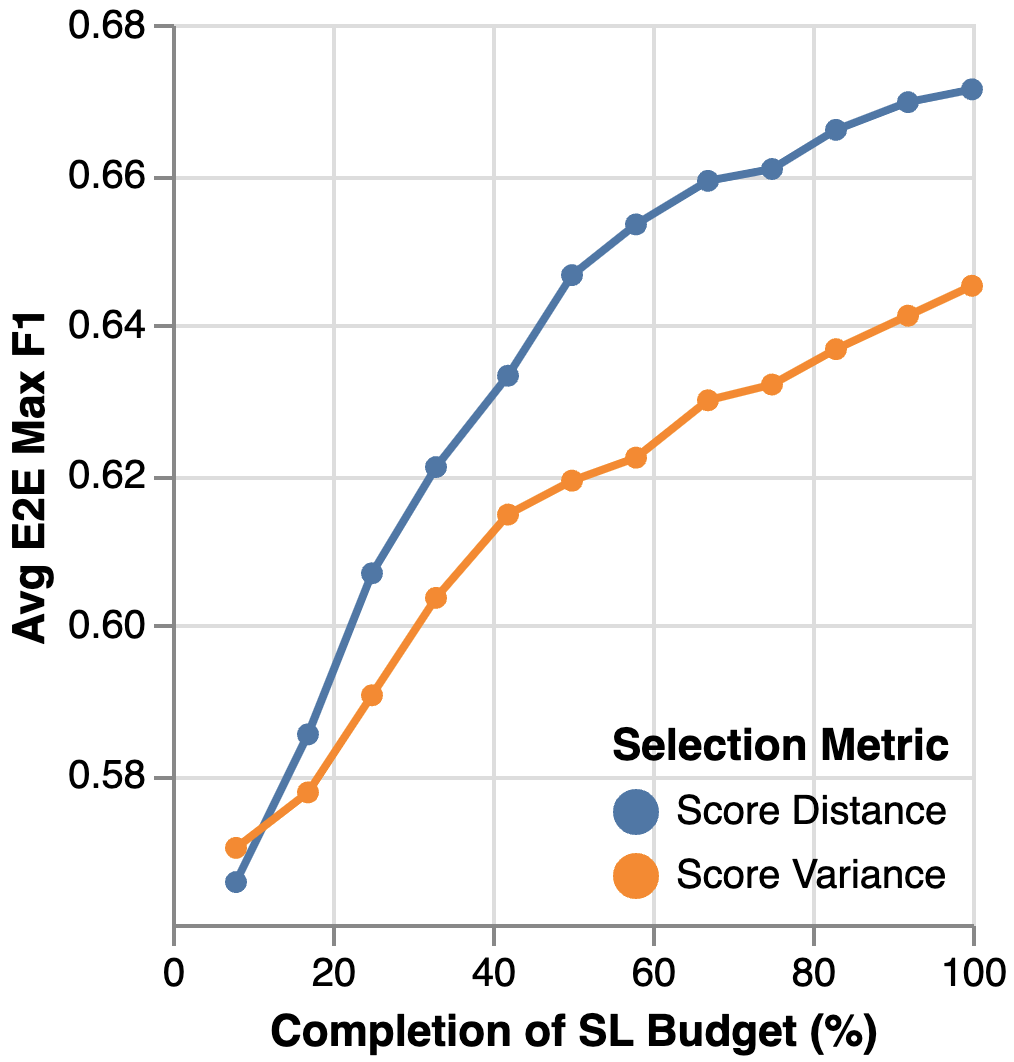}
        \caption{}
        \label{fig:selection_metrics}
    \end{subfigure}~
    \begin{subfigure}[t]{.28\linewidth}
        \includegraphics[width=\linewidth,trim={1.5cm 0 0 0},clip]{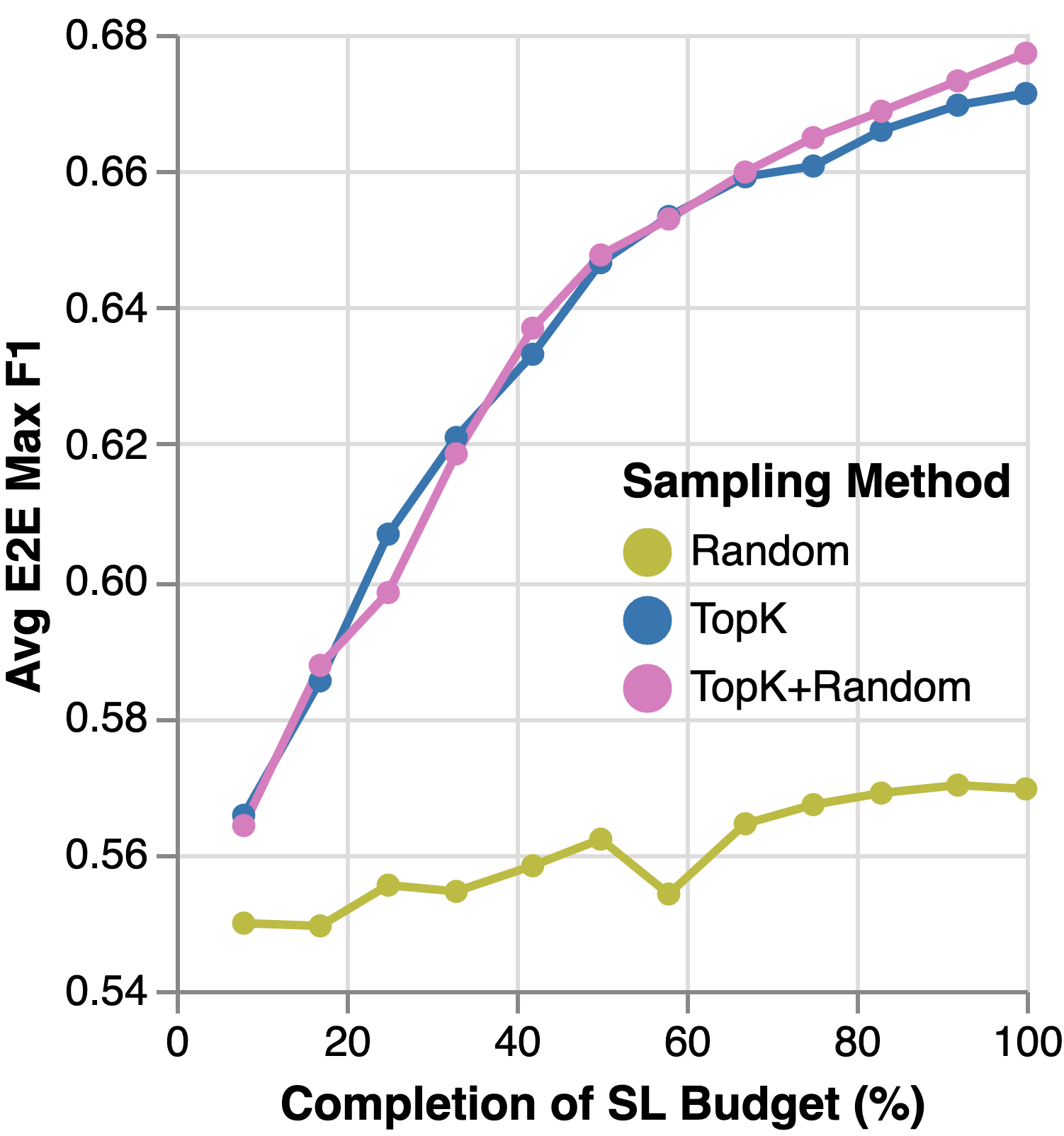}
        \caption{}
        \label{fig:sampling_metrics}
    \end{subfigure}~
    \begin{subfigure}[t]{.28\linewidth}
        \includegraphics[width=\linewidth,trim={1.5cm 0 0 0},clip]{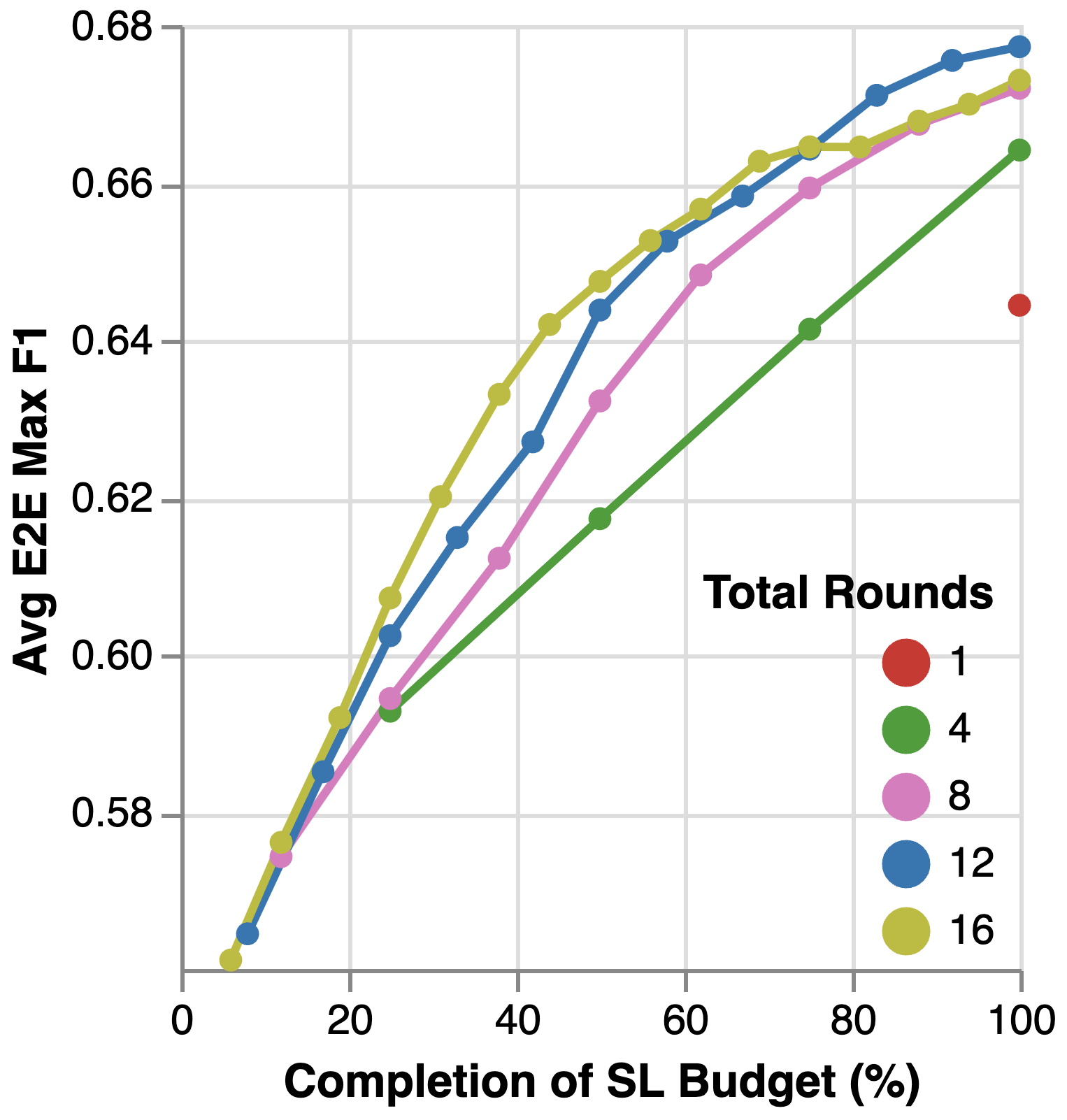}
        \caption{}
        \label{fig:number_rounds}
    \end{subfigure}
\caption{Performance comparisons between (a) selection metrics, (b) sampling approaches, and (c) the rate at which we exhaust the budget through different number of rounds of selective labeling. The x-axis denotes the percentage of the total selective labeling budget consumed.}
\end{figure*}

\subsection{Best Performance on Different Domains}\label{sec:domains}
We train three initial models on a randomly sampled and labeled set of 100 documents for each domain. For example, as shown in Figure~\ref{fig:domains}, the initial model for the \textit{\procurement} domain achieves 0.547 F1 on the test dataset. We fine-tune the initial model on a fully labeled 10k document dataset (i.e., the hidden-label set from Table~\ref{tab:datasets}, in which for the purposes of this analysis we use its true labels), resulting in an F1 score of 0.705. The performance gap between these two models is thus 0.158.

Starting from the same initial model, we apply our best selective labeling strategy (which we discuss in the following sections) to reveal the labels from a subset of candidates that comprises only 10\% of the annotation cost of fully labeling the hidden-label dataset. For the \textit{\procurement} domain, this achieves an F1 score of 0.687, which closes the performance gap by 89\%. Similarly, we close the gap by 88\% and 92\% for the \textit{\banking} and \textit{\mortgage} domains, respectively. This demonstrates that our method can dramatically decrease the annotation cost without sacrificing much performance.

\subsection{Selection Metrics}\label{sec:selection_metrics}
In Figure~\ref{fig:selection_metrics} we plot per-round performance of two selection metrics in the \textit{\procurement} domain given the same set of documents and annotation budget (i.e, 10\% cost) and using the top-$k$ sampling methodology.
We observe that not only is computing score distances as the uncertainty indicator much more computationally efficient than variance-based metrics ($10\times$ faster), but it also significantly outperforms the latter as well. 
As we exhaust the budget over time, the advantage of score distance becomes more obvious.

\subsection{Sampling Methodology}\label{sec:sampling_metrics}
Figure~\ref{fig:sampling_metrics} compares performance across different sampling methodologies. As one might expect, pure random sampling is far worse than any other approach -- we believe the initial model is confident in predicting a large quantity of candidates (especially the negatives), and randomly sampling from them does not obtain much useful knowledge. 

The top-$k$ strategies produce much more impressive results. Furthermore, we observe in later rounds that injecting some diversity via randomness achieves slightly better performance than the vanilla top-$k$ approach. We believe this mimics the aggregation of exploitation (top-$k$) and exploration (random) processes, proven to be beneficial in reinforcement learning applications~\citep{ishii2002control}. This also confirms our suspicion that top-$k$ alone can lead us into selecting many uncertain examples which are in fact very similar to one another.

\subsection{Multi-round Setting}\label{sec:number_rounds}
In Figure~\ref{fig:number_rounds}, we compare 5 learning curves, each of which denotes selecting the same number of candidates in total (10\% annotation cost) over a different number of rounds. For example, the 16-round experiment selects $\frac{1}{16}$ of the total budget in each round, while the 1-round experiment utilizes the entire budget in a single round.

As we increase the total number of rounds, the model tends to yield better extraction performance until it peaks at about 12 rounds. This finer-grained strategy usually performs better than coarser ones but the gains become marginal at a higher number of rounds.
Interestingly, we find that using up just half the budget in the first 8 rounds of a 16-round experiment achieves slightly better performance than exhausting the entire budget in the 1-round experiment. This comparison underscores the importance of employing a multi-round approach.

\begin{table}[t]
    \centering
    \resizebox{.85\linewidth}{!}{
    \begin{tabular}{l|c|c}
    \hline
        Models & Avg E2E Max F1 (std.) & $\Delta$ \\
        \hline
        \hline
        SL & 0.671 (0.006) & - \\
        SL+CS & 0.679 (0.005) & +1.2\%\\
        SL+CC & 0.675 (0.005) & +0.6\%\\
        SL+AIN & 0.683 (0.009) & +1.8\%\\
        SL+CS+CC+AIN & 0.687 (0.005) & +2.1\% \\
        \hline
    \end{tabular}
    }
    \vspace{10pt}
    \caption{Ablation Study. SL denotes selective labeling utilizing the top-$k$ sampling and score distance metric. CS, CC, and AIN represent calibrating scores, capping candidates and automatically inferring negatives.}
    \label{tab:ablation_study}
\end{table}

\subsection{Ablation Study}\label{sec:ablation_study}
Table~\ref{tab:ablation_study} presents an ablation study to understand the impact of different diversity strategies. \textsc{SL} represents a 12-round selective labeling method using top-$k$ sampling on the score distance metric. We separately add one feature at a time to test the effectiveness of calibrating scores (\textsc{CS}), automatically inferring negatives (\textsc{AIN}) and capping candidates (\textsc{CC}). Results show that every feature improves the model, but we achieve the largest improvement when applying all features in \textsc{SL+CS+CC+AIN}. It is reasonable to conclude that increasing diversity intelligently helps us select more useful candidates than relying on the uncertainty metric alone.

\subsection{Initial Labeled Dataset Size}\label{sec:starting_dataset_sizes}
Given the dependence of the selective labeling method on an initially labeled small dataset, it is imperative that we evaluate how the approach is affected by the number of documents in this initial dataset. We experiment with initial datasets of 50, 100, and 250 documents in the \textit{\procurement} domain using our best selective labeling strategy and a budget equivalent of 10\% cost of annotating the ``unlabeled'' dataset.

\begin{figure}[t]
\includegraphics[width=.9\linewidth]{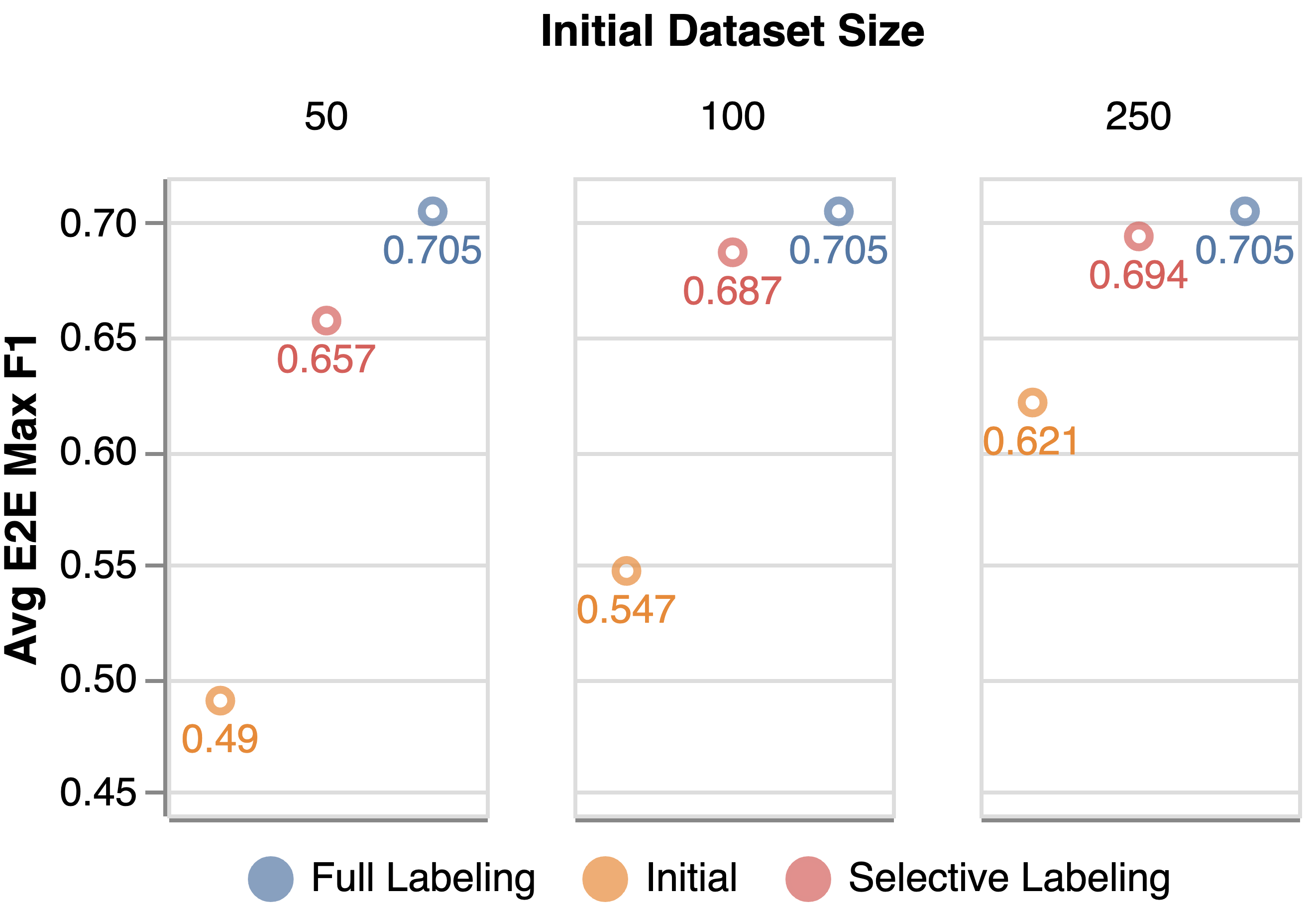}
\centering
\caption{Comparison among three initial dataset sizes in the \textit{\procurement} domain. We present the same three approaches as in Figure \ref{fig:domains}: Initial is trained on the initial dataset alone, Selective Labeling selects the equivalent of 10\% annotation cost in candidates, and Full Labeling fine-tunes from the initial model on the full hidden-label data.}
\label{fig:training_dataset_sizes}
\end{figure}

Figure~\ref{fig:training_dataset_sizes} indicates that the size of the initial dataset greatly impacts the performance of the model trained solely on those initial training sets, but has starkly less of an impact once we apply selective labeling. We close the performance gap by 77\%, 89\%, and 87\%, for initial dataset sizes of 50, 100, and 250, respectively. 
We can conclude that selective labeling is capable of finding useful candidates to significantly improve the model performance even at a cost of only 10\% of the annotation budget. And it is not surprising that the selective labeling gains may suffer when the initial dataset is too small (e.g. 50).

\subsection{Per-field Extraction Analysis}\label{sec:perfield}
\begin{figure*}[t]
\includegraphics[width=.95\linewidth]{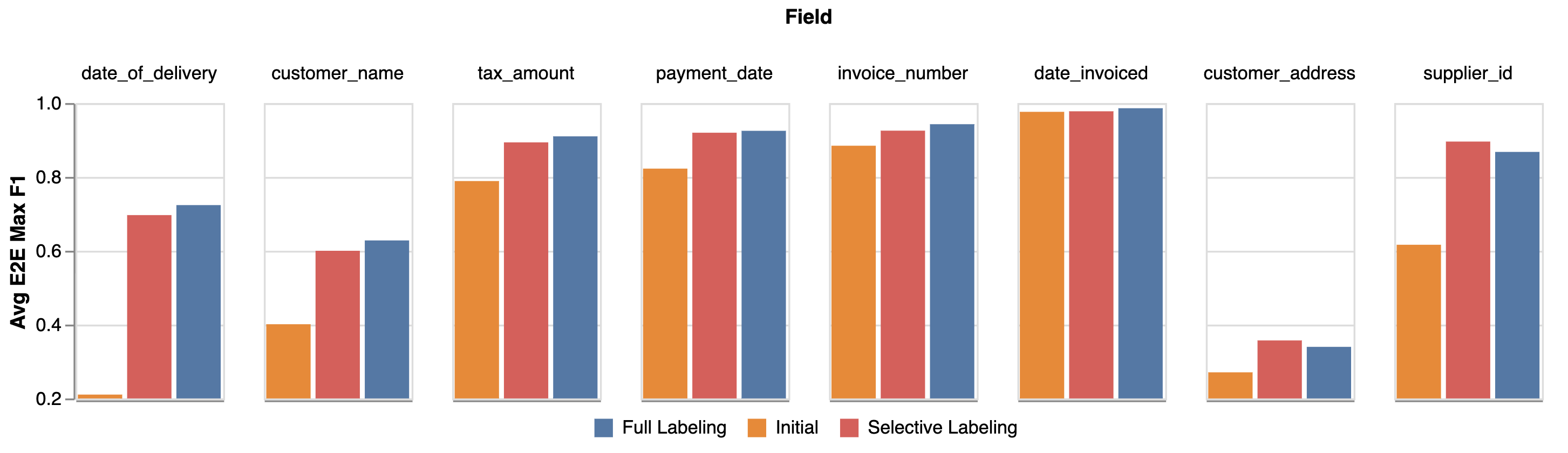}
\centering
\caption{Per-field comparison among Initial, Selective Labeling and Full Labeling. Initial is trained on the initial dataset alone, Selective Labeling selects the equivalent of 10\% annotation cost in candidates, and Full Labeling fine-tunes from the initial model on the full hidden-label data.}
\label{fig:fields}
\end{figure*}

We examine the extraction performances of eight fields from the \textit{Supplier Chain} document type in Figure~\ref{fig:fields} (initial dataset size is 100) to better understand where selective labeling works well.
\red{The recall of candidate generation for these fields varies from 30\% to 99\% showing that selective labeling works even when candidate generation is not perfect.}
We observe that the big gap between Initial and Full Labeling is almost completely closed by selective labeling in fields such as \textit{date\_of\_delivery} and \textit{customer\_name}. Unsurprisingly, the algorithm results in strong improvements for fields where the initial model hasn't seen enough examples.
For frequent fields such as \textit{date\_invoiced} and \textit{invoice\_number}, the initial model performs well, and there is not much room for improvement. Consequently, few candidates are selected and the resulting Selective Labeling model remains competitive on these fields. 

We also notice that Selective Labeling slightly outperforms Full Labeling in some fields such as \textit{supplier\_id} and \textit{customer\_address}. We believe there must exist noisy annotations in the full dataset, and Selective Labeling avoids training the model on some of them. These skipped instances are candidates that our model was certain about, thus including these in the full training dataset likely presented the model with contradictory examples.

\section{Related Work}
\label{sec:relatedwork}
\noindent\textbf{Form Extraction.} There have been numerous recent studies on information extraction for form-like documents. 
Existing approaches either individually categorize every text span in the document~\citep{majumder2020representation,zhang2021integrated} or formulate the task into a sequence modeling problem~\citep{aggarwal2021form2seq,lee2022formnet,garncarek2021lambert,xu-etal-2021-layoutlmv2} and encode texts, layouts, and visual patterns into feature space. 
While these approaches produce state-of-the-art extraction systems, they require large amounts of labeled training data to do so. In our work, we do not propose a new model architecture but instead, focus on the cost of acquiring labeled data for such extraction models.

\vspace{5pt}
\noindent\textbf{Active Learning.} We refer to \citep{settles2009active, fu2013survey, ren2021survey} for an extensive review of the literature. In our work, we are interested in a pool-based selection strategy that assumes a large unlabeled set to select samples from and request for human annotation. 
Two popular approaches for requesting annotation are (1) uncertainty-based selection~\citep{lewis1994sequential} which can measure the uncertainty based on entropy~\citep{ko1995exact}, least confidence~\citep{culotta2005reducing}, or maximum margin~\citep{boser1992training}; and (2) committee-based selection~\citep{seung1992query}, which select instances based on disagreement upon multiple predictions~\citep{gal2016dropout, kirsch2019batchbald, bengar2021deep}. 
Methods that are only concerned with uncertainty might introduce redundancy or skew the model towards that particular area of the distribution. 
Researchers seek to increase the diversity by forcing the selection to cover a more representative set of examples~\citep{yang2017suggestive, yin2017deep,sener2017active} or incorporating discriminative learning to make the labeled set and the unlabeled pool indistinguishable~\citep{gissin2019discriminative}. 

In recent years, deep learning methods have obtained excellent results on various important supervised learning tasks. Researchers have studied combining active learning with deep learning. Most of the advanced strategies such as Coreset~\citep{sener2017active}, Dropout~\citep{gal2016dropout}, Discriminative Active Learning~\citep{gissin2019discriminative} focus on image classification tasks with convolutional neural networks. \citep{zhang2017active, prabhu2019sampling, siddhant2018deep, zhang2021cartography, yuan2020cold} studies deep active learning on NLP tasks such as text classification. Hybrid methods that combine uncertainty and diversity sampling have also been proposed for batch active learning in the deep neural network context~\citep{ash2019deep,yin2017deep,shui2020deep}.

To the best of our knowledge, we are the first to customize active learning strategies to reduce the annotation cost in the form-like document extraction task. In our selective labeling experiments, we explore a variety of informativeness-based selection strategies due to their simplicity and promising performance. We also explore introducing diversity by reallocating a portion of the labeling budget for random sampling as well as through proposing task-aware methods, such as automatic negative inference and capping candidates.

\red{Crowdsourcing is a related line of work, where one target is to simply/minimize questions to annotation crowdworkers~\citep{BoimGMNPT12,zheng2017truth,RoyLTAD15}. Our goal, however, is not to measure or understand the annotator quality but to argue for structuring the labeling task in a way that imposes a low cognitive burden and modify the algorithm to make it cheaper to acquire labeled data. As we explore asking the annotators to do more complex tasks like correcting pseudo-labels or re-drawing bounding boxes, crowdsourcing literature can be a great reference for making efficient annotation assignment.}

\section{Conclusions and Future Work}
\label{sec:conclusion}

We have presented a new approach to acquire labeled data for form extraction tasks that reduces the annotation cost by $10\times$ as compared to fully labeling a large corpus, without sacrificing much extraction performance. The key insight is to transform the annotation task into a ``yes/no'' task and leverage a model type that can be trained on partially labeled documents in a multi-round active learning setting. We proposed novel techniques that take advantage of the characteristics of the problem to further improve extraction performance in the context of our selective labeling strategy. Thus, our approach has the potential to overcome the bottleneck of obtaining large amounts of high-quality training data for hundreds of document types.

There are several future avenues for investigation. First, we simplified the annotation task to a binary ``yes/no'' question. Another approach is to allow the annotator to either accept the candidate annotation, or {\em correct} it -- either by deleting it or by adjusting the bounding box. For certain text fields it can be valuable to adjust spans to include/exclude details like salutations from a name field (``Mr.'', ``Dr.'' etc.)  or names from an address. The cost model for such an option is more complex than ``yes/no'', but can be used to build on the results in this paper.
Second, many recent approaches~\citep{xu-etal-2021-layoutlmv2,lee2022formnet} treat this as a sequence-labeling problem and use a layout-aware language model. \red{This modeling approach is attractive since it doesn't require candidate generators for fields.} Adapting selective labeling to a sequence-labeling model requires tackling several problems: a) getting uncertainty estimates for a given span from a sequence labeling model, b) training a sequence labeling model using partially labeled documents, and c) optionally, eschewing candidate-generators entirely and generating both candidate-spans and their uncertainty estimates form the sequence labeling model. We hope to explore the multiple ways to tackle each of these problems in future work.
Third and finally, the active-learning approaches we used in this paper are relatively simple. Additional efficiency might be available by using more recently developed advanced techniques~\citep{zhang2017active, zhang2021cartography, ash2019deep}.



\bibliographystyle{ACM-Reference-Format}
\bibliography{reference}

\end{document}